\documentclass[graybox]{svmult}

\usepackage{type1cm}        % activate if the above 3 fonts are
\usepackage{makeidx}         % allows index generation
\usepackage{graphicx}        % standard LaTeX graphics tool
\usepackage{multicol}        % used for the two-column index
\usepackage[bottom]{footmisc}% places footnotes at page bottom
\usepackage{natbib}

\usepackage{amsthm}
\usepackage{xcolor}
\usepackage{url}

\usepackage{booktabs}
\usepackage{amssymb}
\usepackage{amsmath}
\usepackage{amsfonts}
\usepackage{mathrsfs}
\usepackage{bbm}
\usepackage{natbib}
\usepackage{comment}
\usepackage{placeins}

\usepackage{amsmath,amsfonts,bm}

\def\eqref#1{equation~\ref{#1}}
\def\1{\bm{1}}

\DeclareMathAlphabet{\mathsfit}{\encodingdefault}{\sfdefault}{m}{sl}
\SetMathAlphabet{\mathsfit}{bold}{\encodingdefault}{\sfdefault}{bx}{n}
\newcommand{\R}{\mathbb{R}}

\newcommand{\KL}{\mathcal{D}_{\mathrm{KL}}}

\usepackage{hyperref}
\usepackage{url}
\usepackage{amsthm}
\usepackage{amssymb}
\usepackage{graphicx}
\usepackage{makecell}
\usepackage{bbm}

\usepackage{newtxtext}       % 
\usepackage[varvw]{newtxmath}       % selects Times Roman as basic font

\newcommand{\abs}[1]{\left\lvert#1\right\rvert}

\newcommand{\pth}[1]{\left( #1 \right) }
\newcommand{\bk}[1]{\left[ #1 \right]}
\newcommand{\curl}[1]{\left\{#1\right\}}

\newcommand{\variance}{\mathbb{V}}

\newcommand{\dd}{\textup{d}}

\newcommand{\PB}{\textup{PB}}

\newcommand{\risk}{R}
\newcommand{\trisk}{\tilde{\risk}}

\newcommand{\proba}{\pi}
\newcommand{\prior}{\proba_{\textup{p}}}

\usepackage{cleveref}

\crefname{assump}{Assumptions}{assumptions}
\crefname{corollary}{Corollary}{corol.}

\makeindex             % used for the subject index
\begin{document}

\title*{Change of measure through the Legendre transform}
\author{Antoine Picard-Weibel\orcidID{0000-0003-0340-1072} and\\ Benjamin Guedj\orcidID{0000-0003-1237-7430}}
\institute{Antoine Picard-Weibel \at Suez, CIRSEE, France \email{apicard.w@gmail.com}
\and Benjamin Guedj \at University College London, United Kingdom \emph{and} Inria, France \email{b.guedj@ucl.ac.uk}}
\maketitle

\abstract{
PAC-Bayes generalisation bounds are derived via change-of-measure inequalities that transfer concentration properties from a reference measure to all posterior measures. The specific choice of change of measure determines the assumptions required on the empirical risk; in particular, the classical Donsker--Varadhan theorem leads to bounds relying on bounded exponential moments. We study change-of-measure inequalities based on \(f\)-divergences, obtained by combining the Legendre transform of \(f\) with the Fenchel--Young inequality. Beyond their intrinsic interest in probability theory, we show how these inequalities are helpful in learning theory and yield PAC-Bayes bounds under tailored assumptions on the empirical risk, thereby extending the range of conditions under which PAC-Bayesian guarantees can be established.}

\section{Introduction}
\label{sec:ch2_1:PAC_Bayes_f_div}

PAC-Bayes theory \citep[see][for recent monographs]{alquier2024user,hellstrom2023generalisation} provides a powerful framework to derive generalisation guarantees for learning algorithms by combining probabilistic reasoning with information-theoretic principles. At its core, it relates the expected risk of a stochastic predictor to its empirical performance, while controlling complexity through a divergence between posterior and prior distributions on predictors. Classical PAC-Bayes bound rely on two key assumptions:
\begin{itemize}
	\item the empirical risk function is expressed as the empirical mean of $n$ loss functions $\ell_i$ of the form $\ell_i(\theta) = \ell(\theta, z_i)$, with independent, identically distributed (i.i.d.) observations $z_i$,
	\item  the loss function $\theta, z\mapsto \ell(\theta, z)$ is positive and bounded.
\end{itemize}
These assumptions are natural in standard supervised learning settings. For instance, in classification, the $0$--$1$ loss is both bounded and compatible with the i.i.d. sampling assumption. Similarly, in regression with bounded outputs and predictors, commonly used losses such as the mean squared error or mean absolute error satisfy these conditions.

However, many modern applications fall outside this classical regime. In particular, when dealing with time series data or unbounded observations, the independence assumption is violated and losses such as the mean squared error are no longer bounded. This setting is prevalent in a wide range of domains, from signal processing to finance, and calls for an extension of the PAC-Bayes framework. Prior work has explored these directions, notably for dependent data \cite{ralaivola2010a,alquier2012model,alquier2018simpler,haddouche2022online} and for unbounded or heavy-tailed losses \cite{catoni2004b,begin2016pac,alquier2018simpler,haddouche2021pac}.

A key obstacle in extending PAC-Bayes bounds lies in the structure of their proofs. Classical results rely on change-of-measure inequalities, such as the Csiszár--Donsker--Varadhan variational formula, which involve exponential moments of the generalisation gap, or of a suitable transform thereof. Consequently, the classical Kullback-Leibler-based route naturally leads to exponential integrability requirements. While these assumptions are compatible with bounded losses, they may be too strong in the presence of unbounded or heavy-tailed losses.

The unbounded loss setting has been investigated in several works: notably, \cite{begin2016pac,alquier2018simpler} go beyond the usual Kullback-Leibler penalisation by considering more general divergences. This shift is not merely technical: it reflects a fundamental link between the divergence used to compare prior and posterior distributions, and the moment assumptions required on the generalisation gap. This perspective has also been explored through alternative change-of-measure inequalities based on $f$-divergences, notably in \cite{ohnishi2021}, where PAC-Bayes bounds are derived under general $f$-divergence penalisation, further emphasising the interplay between divergence choice and moment assumptions. For the power-type $f$-divergences considered in \cite{alquier2018simpler}, this link appears through Hölder's inequality, yielding bounds involving conjugate finite moment rather than exponential moments.

In this paper, we show that this phenomenon is not specific to power divergences. We establish a systematic connection between $f$-divergences and moment assumptions through convex duality. We derive change-of-measure inequalities by combining the Legendre transform of a convex generator $f$ with the Fenchel--Young inequality. This yields PAC-Bayes objectives penalised by arbitrary $f$-divergences and involving the corresponding $f^*$-moment of the generalisation gap, where $f^*$ denotes the Legendre transform of $f$. The resulting framework recovers both the exponential moment condition associated with the Kullback-Leibler divergence and the finite-moment conditions arising from previously studied power divergences.

We then study the tighthness and optimisation of these inequalities. Since several convex generators define the same $f$-divergence up to an affine correction, the induced moment term contains a positional degree of freedom. We show that optimisation over this scalar parameter recovers the exact Legendre transform of the $f$-divergence under mild regularity assumptions. We also introduce a scale parameter, interpretable as a temperature, which provides an additional degree of flexibility. For several tractable divergences, we derive explicit change-of-measure inequalities and discuss the corresponding trade-off between the strength of the divergence penaly and the moment assumptions imposed on the generalisation gap. Finally, we show how these inequalities can be used to construct PAC-Bayes bounds under tailored assumptions on the loss.

\section{Notation}

For a measurable space $(\mathcal{A}, \Sigma_\mathcal{A})$, we denote
\begin{itemize}
    \item $\Pi_\mathcal{A}$ the set of all probability measures on $(\mathcal{A}, \Sigma_\mathcal{A})$,
    \item $\mathcal{F}_\mathcal{A}$ the set of all real-valued measurable functions on $\mathcal{A}$ (with respect to the Borel $\sigma$-algebra on $\mathbb{R}$).
\end{itemize}
When the underlying measurable space is clear from the context, the subscript $\mathcal{A}$ is omitted. For any probability measure $\pi \in \Pi$, we use the following notation:
\begin{itemize}
    \item $\mathrm{L}^1(\pi)$ denotes the set of real-valued measurable functions whose absolute value is integrable with respect to $\pi$;
    \item for any $D \in \mathrm{L}^1(\pi)$, we write
    \[
    \pi[D] := \int D(\gamma)\, \pi(\mathrm{d}\gamma).
    \]
    When $D$ is measurable but not $\pi$-integrable, we allow $\pi[D]$ to take infinite values. When used in inequalities, these values are interpreted in the extended real line so that the inequality remains valid;
    \item $\delta_\gamma$ denotes the Dirac probability measure at point $\gamma$.
\end{itemize}
For two probability measures $\pi_1, \pi_2$ defined on the same measurable space, $\pi_1 \ll \pi_2$ denotes that $\pi_1$ is absolutely continuous with respect to $\pi_2$. The Kullback--Leibler divergence (KL) is defined as
\begin{equation}
\label{eq:ch1:KL_definition}
\KL(\pi_1,\pi_2) =
\begin{cases}
\displaystyle \pi_2\!\left[\frac{\mathrm{d}\pi_1}{\mathrm{d}\pi_2} \log\!\left(\frac{\mathrm{d}\pi_1}{\mathrm{d}\pi_2}\right)\right]
& \text{if } \pi_1 \ll \pi_2, \\
+\infty & \text{otherwise.}
\end{cases}
\end{equation}
Let $\mathcal{A}$ and $\mathcal{B}$ be vector spaces equipped with a bilinear form $(a,b) \mapsto a \cdot b$. The Legendre transform of a function $f : \mathcal{A} \to \mathbb{R}$ is the function $f^* : \mathcal{B} \to \mathbb{R}$ defined by
\[
f^*(b) := \sup_{a \in \mathcal{A}} \left\{ a \cdot b - f(a) \right\}.
\]
For PAC-Bayes applications, we use the following notation. The observed data is denoted $z \in \mathcal{Z}$ and is assumed to be drawn from an unknown probability distribution $\mathbb{P}$. We let $\Gamma$ denote the hypothesis space (or predictor space).

For any predictor $\gamma \in \Gamma$ and observation $z$, the loss is denoted $\ell(\gamma, z)$. The empirical risk is defined as
\[
R(\gamma) := \ell(\gamma, z),
\]
where the dependence on the data is implicit. The true risk is defined as
\[
\trisk(\gamma) := \mathbb{P}[\ell(\gamma, \cdot)].
\]
A PAC-Bayes bound (denoted $\PB$) is a function of a posterior distribution $\pi \in \Pi$, the empirical risk $R$, a prior $\prior \in \Pi$, and a confidence level $\delta \in (0,1)$ such that, under suitable assumptions on $\mathbb{P}$ and $R$, the following holds:
\[
\forall \prior,\ \forall \delta,\qquad
\mathbb{P}\!\left(
\forall \pi,\ 
\pi[\trisk] \leq \PB(\pi, R, \prior, \delta)
\right) \geq 1 - \delta.
\]

\section{From Legendre Transforms to PAC-Bayes objectives}
\label{sec:ch2:Legendre_transform_to_PB_obj}

The Csiszár--Donsker--Varadhan change-of-measure inequality \citep{csiszar1975divergence,donsker1975large}, valid for all bounded measurable functions $D$, states that
\begin{equation}
\label{eq:ch2_1:change_measure}
\log\pth{\prior\bk{\exp D}}= \sup_{\pi\in\Pi}\curl{\pi\bk{D} - \KL(\pi, \prior)}.
\end{equation}
with the convention that $\infty -\infty = -\infty$.

The space $\Pi = \Pi_{\mathcal{H}}$ is an affine subset of the vector space of bounded signed measures $\overline{\mathcal{M}}_{\mathcal{H}}$. The continuous dual of this space is the set of bounded measurable functions \citep{Fichtenholz1934}. The functional $\overline{\KL}:\pi \mapsto \KL(\pi,\prior)$ is convex on $\Pi$. By extending it to $+\infty$ outside of $\Pi$ in $\overline{\mathcal{M}}_{\mathcal{H}}$, it remains convex on the whole space. Since the mapping $(\pi,D) \mapsto \pi[D]$ extends the canonical bilinear form (see \cite{Hildebrandt1934}, Section IV.5, Theorem 1), it follows that
\[
D \mapsto \sup_{\pi\in\Pi}\curl{\pi\bk{D} - \KL(\pi, \prior)}
\]
is the Legendre transform of $\overline{\KL}$, denoted $\overline{\KL}^*$. Hence, the Csiszár--Donsker--Varadhan identity can be written compactly as
\[
\overline{\KL}^*(D) = \log\pth{\prior\bk{\exp D}}.
\]

By construction, Legendre transforms satisfy Fenchel--Young's inequality:
\begin{equation}
\label{eq:ch2_1:Fenchel_Young_KL}
\pi\bk{D}\leq \overline{\KL}^*(D) + \overline{\KL}(\pi).
\end{equation}
This decomposes the bilinear form $(\pi,D) \mapsto \pi[D]$ into a term depending only on $D$ and a term depending only on $\pi$. This property is central for constructing PAC-Bayes bounds. Here, $D$ should be understood as a generalised form of the generalisation gap $\trisk - \risk$. If one can control $\overline{\KL}^*(D)$ using a concentration inequality, then one obtains a uniform control of the average generalisation gap over \emph{all} posterior distributions $\pi$, since $\overline{\KL}(\pi)$ no longer involves the data. This strategy underlies classical PAC-Bayes bounds such as those in \cite{catoni2004b,maurer2004,langford2001bounds,germain2009a}.

Importantly, this approach does not rely on the specific form of the penalisation. For any real-valued functional $P:\Pi\to\mathbb{R}$, one can define its convex conjugate as
\begin{equation}
\label{eq:ch2_1:General_Legendre_definition}
P^*(D) := \sup_{\pi\in\Pi}\left\{\pi[D] - P(\pi)\right\}.
\end{equation}
This yields the Fenchel--Young inequality
\begin{equation}
\label{eq:ch2_1:Fenchel_Young}
\pi\bk{D}\leq P^*(D) + P(\pi)
\end{equation}
for all probability measures $\pi$. This remains valid for any measurable function $D$, with the convention that $\pi[D] = -\infty$ whenever $D \notin \mathrm{L}^1(\pi)$. Moreover, the inequality still holds if $P^*$ is replaced by any upper bound $\overline{P}^*$.

This provides a general route to constructing PAC-Bayes bounds. For instance, taking $D = \trisk - \risk$ in \Cref{eq:ch2_1:Fenchel_Young} yields
\begin{equation}
\label{eq:PAC_Bayes_from_Fenchel_Young}
    \PB(\pi, \risk, \prior, \delta) = \pi[\risk] + Q_\delta + P(\pi),
\end{equation}
where $Q_\delta$ denotes the $(1-\delta)$-quantile of the random variable $\overline{P}^*(\trisk - \risk)$, which does not depend on the posterior $\pi$. The PAC-Bayes learning strategy then consists in minimising the right-hand side over $\pi$. This can be interpreted as a penalised learning problem, where $\pi[\risk]$ measures empirical performance and $P(\pi)$ acts as a complexity penalty. The tightness of the resulting bound depends on the quantile term $Q_\delta$, which in turn depends on how tightly $\overline{P}^*$ can be controlled. This provides a strong incentive to design penalisation functionals $P$ for which sharp upper bounds on $P^*$ can be obtained.

\section{Legendre transform of the \ensuremath{f}-divergence}

\subsection{An initial upper bound}

\label{sec:ch2:UpBound_f_div_Legendre}

For a convex function $f:\R_+ \to \R$ such that $f(1) = 0$, the $f$-divergence $\mathcal{D}_f$ on probability measures on $\mathcal{H}$ is defined as
\begin{equation}
\mathcal{D}_f(\pi_1,\pi_2) = \begin{cases}
\pi_2\bk{f\pth{\frac{\mathrm{d}\pi_1}{\mathrm{d}\pi_2}}} & \pi_1\ll\pi_2 \text{ and } \pi_2\bk{\abs{f\pth{\frac{\mathrm{d}\pi_1}{\mathrm{d}\pi_2}}}}<\infty,\\
+\infty & \textup{otherwise}.
\end{cases}
\end{equation}

Since $f$ is convex, Jensen's inequality implies that $\mathcal{D}_f(\pi_1, \pi_2) \geq f(1) = 0$. Moreover, $\mathcal{D}_f(\pi,\pi) = \pi\bk{f(1)} = 0$. Hence, $\mathcal{D}_f$ defines a notion of proximity between measures; since it is not symmetric and may fail to satisfy the triangle inequality, it is not a distance but only a divergence.

The most popular $f$-divergence is without doubt the KL divergence, obtained for the function $f(x) = x\log(x)$. We remark that the $f$-divergence $\mathcal{D}_f$ does not uniquely identify the convex function $f$: there exist multiple convex functions $f_1, f_2$ such that $\mathcal{D}_{f_1} = \mathcal{D}_{f_2}$. These are related by a simple relationship, given in the following lemma.

\begin{lemma}
\label{Lemma:identification}
Let $f_1$ and $f_2$ be two convex functions on $\mathbb{R}_+$ such that $f_1(1) = f_2(1) = 0$. 
Let $(\mathcal{H}, \Sigma_\mathcal{H})$ be a measurable space such that $\abs{\Sigma_\mathcal{H}}>2$. Then the $f$-divergence operators defined on the probability measures on $(\mathcal{H}, \Sigma_\mathcal{H})$ by $f_1$ and $f_2$ are equal if and only if there exists $c\in\R$ such that $\forall x\in \R_+$, $f_1(x) = f_2(x) + c(x-1)$.
\end{lemma}

\begin{proof}[Proof of \Cref{Lemma:identification}]
The mapping $f \mapsto \mathcal{D}_f$ is linear. Hence solving $\mathcal{D}_{f_1}=\mathcal{D}_{f_2}$ amounts to solving $\mathcal{D}_f = 0$ (under the constraint $f(1)=0$, although this condition can be relaxed).

If $\abs{\Sigma_\mathcal{H}}>2$, there exists a set $B\in\Sigma_\mathcal{H}$ such that $B\neq \varnothing$ and $B\neq \mathcal{H}$. Hence we can consider $h_1$, $h_2$ such that $h_1\in B$, $h_2\in B^c$. We can then define the Dirac probability measures $\delta_{h_1}$ and $\delta_{h_2}$ as $\delta_{h_1}[S] = \mathbbm{1}_{h_1\in S}$ and $\delta_{h_2}[S] = \mathbbm{1}_{h_2\in S}$.

Then, using the measures $\pi_a = a\delta_{h_1} + (1-a)\delta_{h_2}$ and $\pi_b = b\delta_{h_1} + (1-b)\delta_{h_2}$, one obtains, for all $0<a<1$ and $0<b<1$,
\begin{align*}
\mathcal{D}_f(\pi_a, \pi_b) = b f\pth{\frac{a}{b}} + (1-b)f\pth{\frac{1-a}{1-b}} = 0.
\end{align*}

Consider the case where $a\neq b$, and define $x = \frac{a}{b}$ and $y = \frac{1-a}{1-b}$. By inverting this system, one obtains $b = \frac{y-1}{y-x}$ and $a = x\frac{y-1}{y-x}$. Note that for all $0<x<1<y$, this yields $0<b<1$ and $0<a<1$. Hence, for all $0<x<1<y$, $f$ satisfies
\begin{align*}
(y-1)f(x) + (1-x)f(y) = 0.
\end{align*}

Fixing $y=2$ implies that for all $0<x<1$, $f(x) = (x-1)f(2)$. In particular, $f(1/2) = -\frac{1}{2}f(2)$. Setting now $x=1/2$, for all $y>1$, we obtain $f(y) = -2(y-1)f(1/2) = (y-1)f(2)$. Hence for all $x\neq 1$, $f(x) = (x-1)f(2)$. Finally, using $\mathcal{D}_f(\pi,\pi)=0$, it follows that $f(1)=0$, which completes the proof\footnote{We have implicitly assumed that singletons $\{h\}$ for $h \in\mathcal{H}$ are measurable. This assumption can be relaxed to $\abs{\Sigma_\mathcal{H}}>2$ (i.e., the $\sigma$-algebra is not limited to the empty set and the whole space), by considering a set $\overline{h}\not\in\{\varnothing, \mathcal{H}\}$, letting $h_1 = \overline{h}$ and $h_2 = \overline{h}^c$, and defining $\delta_{h_1}$ and $\delta_{h_2}$ as any probability measures supported on these sets.}.
\end{proof}

We will show later that this lack of uniqueness has a significant impact on change-of-measure inequalities.

Let us now consider the $f$-divergence induced by a function $f$. Since $f$ is convex, it admits a convex conjugate $f^*$, defined for all $y\in\R$ by
\begin{equation}
\label{eq:ch2_1:legendre_f}
f^*(y) = \sup_{x\geq 0} xy - f(x).
\end{equation}

The pair $(f,f^*)$ satisfies Fenchel--Young's inequality, that is, for all $x,y$,
\begin{equation}
xy \leq f(x) + f^*(y).
\end{equation}

This implies that for all $\pi\ll\prior$ and all measurable real-valued functions $D$,
\begin{align*}
\pi\bk{D} &= \prior\bk{D\frac{\mathrm{d}\pi}{\mathrm{d}\prior}} \\
&\leq \prior\bk{f^*\circ D + f\pth{\frac{\mathrm{d}\pi}{\mathrm{d}\prior}}} \\
&\leq \prior\bk{f^*\circ D} + \mathcal{D}_f(\pi,\prior).
\end{align*}

This is a Fenchel--Young inequality of the form \Cref{eq:ch2_1:Fenchel_Young}. As a consequence, we obtain the following upper bound on the Legendre transform of $\overline{\mathcal{D}_f}$:
\begin{equation}
\label{Eq:raw_bound}
\overline{\mathcal{D}_f}^*(D)\leq \prior\bk{f^*\circ D}.
\end{equation}

\begin{remark}[Some properties of Legendre transforms]\label{remark:ch2_1:f_star_properties}
We list here some useful properties of $f^*$, valid for any convex function $f$ such that $f(1) = 0$. We denote by $\partial f(x)$ the subdifferential of $f$ at $x$.

\begin{itemize}
\item $\forall x$, $f^*(x) \geq x$;
\item define $f'(0) := \inf \bigcup_{t>0}\partial f(t)$. If $f'(0)> -\infty$, then $f^*$ is constant on $]-\infty, f'(0)]$ and takes value $f^*(f'(0))$. As a consequence, $\partial f^*(x) = \{0\}$ on the interval $]-\infty, f'(0)[$;
\item $\inf_{x\in\R}\bigcup\partial f^* = 0$. As a consequence, $f^*$ is non-decreasing;
\item if $f^*$ is differentiable, then $f^*(t) = t{f^*}'(t) - f\circ {f^*}'(t)$;
\item define $f'(\infty) = \sup \bigcup_{t>0} \partial f(t)$. If $f'(\infty)< \infty$, then $\forall x > f'(\infty)$, $f^*(x) = \infty$.
\end{itemize}

A consequence of the first and third properties is that, for lower-bounded $D$, the condition $\prior\bk{D}<\infty$ is necessary for $\prior\bk{f^*(D)}<\infty$.

A consequence of the second property is that if $f'(0) > - \infty$, the moment condition on $D$ involves a threshold on its values.

A consequence of the last property is that Fenchel--Young's inequality becomes trivial whenever $D$ is not bounded by $f'(\infty)$. In other words, penalisation with $f$-divergences such that $\liminf f(x)/x <\infty$ leads to the requirement that $D$ is $\prior$-almost surely bounded (otherwise, the change-of-measure inequality is trivial).
\end{remark}

\subsection{Refinement of the upper bound}

\label{sec:ch2:UpBound_Legendre_refinement}

Let us evaluate the upper bound given by \Cref{Eq:raw_bound} in the case of the KL divergence. As stated above, the KL divergence is an $f$-divergence for $f(x)= x\log(x)$. The Legendre transform of this function is $f^*(x) = \exp(x-1)$. As a result, we obtain the upper bound
\begin{equation*}
\overline{\KL}^*(D) \leq \exp(-1)\prior\bk{\exp(D)}.
\end{equation*}
This is a very loose upper bound: the bound we have just constructed is \emph{exponentially} larger than the true value. The result so far is therefore too loose to be usable in practice, as it leads to poor rates in the confidence level.

As noted above, several convex functions $f$ define the same $f$-divergence. However, since the map from $f$ to $f^*$ is one-to-one for proper convex functions through biconjugation, these different choices of $f$ do not define the same moment term $f^*$. This offers a degree of freedom over which our bound can be minimised. Denoting by $f_c$ the function $x \mapsto f(x) + c(x-1)$, the convex conjugate of $f_c$ can be inferred from the convex conjugate of $f$ and is given by $f_c^*(x) = f^*(x-c) + c$ (to see this, notice that $xy - f_c(x) = c + (x(y-c) - f(x))$, and that the supremum of the second term is by definition $f^*(y-c)$). Hence we can deduce that
\begin{equation}
\label{eq:ch2_1:bound_legendre_inf_c}
\overline{\mathcal{D}_f}^*(D) \leq\inf_{c\in\R}\prior\bk{f^*(D -c)} + c.
\end{equation}

Does this bridge the gap between the Kullback--Leibler Legendre transform and its upper bound? In this case, it is easy to optimise over the positional parameter $c$, since the bound becomes $\exp(-1-c)\prior\bk{\exp(D)} + c$. The minimum is obtained for $c = \log(\prior\bk{\exp(D)}) -1$ and exactly recovers the Csiszár--Donsker--Varadhan formula. Hence minimisation over a single parameter allows us to move from a very loose bound to the tightest additive bound achievable, since for all bounded $D$, there exists a probability measure $\pi$ for which the inequality becomes an equality.

We remark that the same degree of freedom could have been obtained by replacing $D$ by the function $D-c$. An upper bound on $\pi[D-c]$ translates into an upper bound on $\pi[D]$ by adding $c$ to both sides.

One may ask whether this supplementary degree of freedom is sufficient to recover the exact Legendre transform of a generic $f$-divergence. We show in the following theorem that this is the case under mild assumptions.\medskip

\begin{theorem}[Legendre transform of $f$-divergences]
\label{thm:ch2_1:theorem_optim_c}
Consider a differentiable convex function $f:\R_+\to\R$ such that $f(1) = 0$ and a prior measure $\prior$ on $\mathcal{H}$.

Then for any positive measurable function $D$,
\begin{equation}
\label{eq:ch2:loose_bound}
\overline{\mathcal{D}_f}^*(D) := \sup_{\pi\ll\prior}\curl{ \pi[D]- \mathcal{D}_f(\pi,\prior)}\leq \inf_{c\in \R} \prior\left[f^*\left(D-c\right)\right] + c.
\end{equation}

If $f$ is such that $f^*$ is differentiable, with continuous derivative, and if $D$ is such that there exists $c_1\in\R$ satisfying $1\leq\prior\bk{{f^*}'(D-c_1)} <\infty$, then
\begin{equation}
\label{eq:ch2_1:minimisation_on_c}
\overline{\mathcal{D}_{f}}^*(D)=\inf_{c\in \R} \prior\left[f^*\left(D-c\right)\right] + c.
\end{equation}
Moreover, any $c^*$ such that $\prior\left[{f^*}'(D-c^*)\right]=1$ is a minimiser of the right-hand side of \Cref{eq:ch2_1:minimisation_on_c}, while the probability measure defined by $\frac{\mathrm{d}\pi^*}{\mathrm{d}\prior} = {f^*}'(D-c^*)$ maximises \Cref{eq:ch2_1:General_Legendre_definition}.
\end{theorem}

\begin{proof}[Proof of \Cref{thm:ch2_1:theorem_optim_c}]

Starting from \Cref{eq:ch2_1:bound_legendre_inf_c}, we know that
\begin{equation}
\label{eq:ch2_1:sup_to_inf}
\sup_{\pi\in\Pi} \left\{\pi\left[D\right]- \mathcal{D}_f(\pi,\prior)\right\}=\overline{\mathcal{D}_{f}}^*(D)\leq \inf_{c\in\R}\left\{\prior \left[f^*(D-c)\right]+c\right\}.
\end{equation}
Let us assume for the moment that there exists $c^*\in\R$ such that
$\prior\left[{f^*}'(D-c^*)\right]=1$. Since ${f^*}'$ takes non-negative values, we can define the probability measure $\pi^*$ such that $\frac{\mathrm{d}\pi^*}{\mathrm{d}\prior}={f^*}'(D-c^*)$. Therefore
\begin{align*}
\overline{\mathcal{D}_{f}}^*(D)&=\sup_{\pi\in\Pi} \pth{\pi\bk{D}- \mathcal{D}_f(\pi,\prior)}\\
&\geq \prior\bk{D{f^*}'(D-c^*) - f\circ{f^*}'(D-c^*)}\\
&= \prior\bk{(D-c^*){f^*}'(D-c^*) - f\circ{f^*}'(D-c^*)} + c^*\\
&= \prior\bk{f^*(D-c^*)}+c^* \\
&\geq\inf_{c} \curl{\prior\bk{f^*(D-c)}+c},
\end{align*}
where the first inequality is obtained by considering $\pi^*$, the second equality uses the identity $\prior\bk{{f^*}'(D-c^*)}=1$, and the third equality uses $x {f^*}'(x) - f({f^*}'(x)) = f^*(x)$. This inequality, combined with \Cref{eq:ch2_1:sup_to_inf}, implies \Cref{eq:ch2_1:minimisation_on_c}. Moreover, for any $c^*$ satisfying $\prior[{f^*}'(D- c^*)] = 1$, it follows from the sequence of inequalities above that
\[
\overline{\mathcal{D}_{f}}^*(D)\geq \prior\bk{f^*(D-c^*)}+c^*  \geq\inf_{c} \curl{\prior\bk{f^*(D-c)}+c} = \overline{\mathcal{D}_{f}}^*(D),
\]
and hence that $\overline{\mathcal{D}_{f}}^*(D) = \prior\bk{f^*(D-c^*)}+c^*$.

It remains only to prove the existence of $c^*$. Define $M:c\to \prior\bk{{f^*}'(D-c)}$. Our assumptions guarantee that there exists $c_1$ such that $M(c_1) \geq 1$. If $M(c_1)=1$, this concludes the proof. If $M(c_1)>1$, notice that $M$ is non-increasing, and for all $c>c_1$, it is bounded by $M(c_1)<\infty$. Since ${f^*}'(x)\rightarrow_{x\rightarrow-\infty} 0$, it follows from Lebesgue's dominated convergence theorem that $M(c)\rightarrow_{c\rightarrow\infty} 0$. Moreover, since ${f^*}'$ is continuous, it follows that $M(c)$ is continuous on $]c_1, \infty[$ \citep{Schilling2005}. Hence, by the intermediate value theorem, there exists $c^*> c_1$ such that $M(c^*)=1$.
\end{proof}

\begin{remark}[Some intuition on the proof of \protect \Cref{thm:ch2_1:theorem_optim_c}]\label{rmk:ch2:theorem_optim_c_through_Lagrange}
The proof of \Cref{thm:ch2_1:theorem_optim_c} is based on the form of the maximiser in the definition of the Legendre transform. It is possible to motivate this form by considering a Lagrange multiplier. Starting from the definition of the Legendre transform, one can reframe the optimisation problem over the probability measure $\pi$ as an optimisation problem over a positive function $g$, with criterion
\begin{align*}
\sup_{c\in\R}L(c, g) = \prior\bk{D g - f\circ g} - c (\prior\bk{g} -1) 
\end{align*}
where $c$ is a Lagrange multiplier. Considering a perturbation function $\delta g$, one obtains for $\varepsilon\rightarrow 0$
\begin{align*}
L(c, g + \epsilon \delta g) - L(c, g) = \epsilon (\prior\bk{ (D- f'\circ g - c )\delta g}) + o(\epsilon).
\end{align*}
This implies that, at the optimum $g^*$, $D - c = f'\circ g^*$ wherever $g^*>0$, and $D-c - f'\circ g^* \geq0$ wherever $g^* = 0$. This implies that $g^* = {f^*}' \circ (D-c)$ for all points. Moreover, the Lagrange multiplier must be such that $\prior\bk{g^*} = 1$, hence that $\prior\bk{{f^*}'(D-c)} = 1$. For such a $c^*$, the value of the objective is
\begin{align*}
\prior\bk{D g^* - f\circ g^*}
&= \prior\bk{D {f^*}'\circ(D-c^*) - f\circ {f^*}'\circ (D-c^*)} \\
&= \prior\bk{(D-c^*) {f^*}'\circ(D-c^*) - f\circ {f^*}'\circ (D-c^*)} + c^*\\
&= \prior\bk{f^*(D-c^*)} + c^*.
\end{align*}
This motivates the value of $c$ achieving the lower bound.
\end{remark}

\subsection{Further improvement for regular \ensuremath{f}}

\label{sec:ch2:UpBound_regular_f}
The optimisation problem involved in \Cref{eq:ch2_1:bound_legendre_inf_c} is not always practicable, and it might be necessary to use the bound given by some approximation of $c^*$. This motivates the search for tighter bounds of $\overline{D}_f^*$ at a given $c$, that is to say functions $\tilde{D}_{f,c}^*$ such that 
$$\overline{\mathcal{D}}_f^*(D) \leq \tilde{\mathcal{D}}_{f,c}^*(D)\leq \prior\bk{f^*(D-c)}+c.$$

We show that when $f$ is twice differentiable and that its second derivative is such that $1/f''$ is concave, tighter bounds can be constructed.

\begin{theorem}[A tighter upper-bound for regular $f$]
\label{thm:ch2_1:tighter_approx}
Consider a twice differentiable convex function $f:\R^+\to \R$, such that $f(1)=0$ and $1/f''$ is concave. Then, denoting 

\begin{equation*}
 \tilde{\Delta}_f(D) :=
 \begin{cases}
 f^*\circ f'\left(\prior\bk{{f^*}'\circ D}\right) - f'(\prior\bk{{f^*}'\circ D})& \text{if } \prior\bk{\abs{{f^*}'\circ D}} < \infty\\
 0 & \text{else},
\end{cases}
\end{equation*}
one can upper bound the Legendre transform of the f-divergence for $D$ lower bounded by
\begin{equation}
\label{eq:ch2:tight_upper_bound}
\overline{\mathcal{D}}_{f}^*(D)
\leq \inf_{c\in\R}\prior\bk{f^*\circ (D-c)} - \tilde{\Delta}_f(D-c) + c.
\end{equation}
\end{theorem}

\begin{proof}
Let us remark that upper bounding $\mathcal{D}_f^*(D)$ by a function $G(D)$ is equivalent to establishing a Fenchel--Young inequality where $\mathcal{D}_f^*(D)$ is replaced by $G(D)$, since 

\begin{equation*}
G(D) + \mathcal{D}_f(\pi, \prior) \geq \pi[D] \Rightarrow G(D) \geq \sup_{\pi} \pi[D] - \mathcal{D}_f(\pi, \prior) = \mathcal{D}_f^*(D).
\end{equation*}
We will thus prove a Fenchel--Young version of \Cref{eq:ch2:tight_upper_bound} for $c=0$. Replacing $D$ by $D-c$ in the resulting Fenchel--Young inequality the implies the inequality for all $c$.

The proof starts with Lemma 14.2 in \cite{boucheron2013concentration}, which states that for any $f$ convex, twice differentiable on $\R_+^*$ such that $\frac{1}{f''}$ is concave, for any $Z>0$ such that $f(Z)$ is $\prior$-integrable, then 
\begin{equation*}
\prior\bk{f(Z)}- f\pth{\prior\bk{Z}}= \sup_{T\neq 0}\curl{\prior\bk{\pth{ f'(T)-f'(\prior\bk{ T})}\left(Z-T\right) + f(T)}- f\left(\prior\bk{ T}\right)}
\end{equation*}
where the supremum is taken on all non negative $\prior$-integrable random variables $T$. The maximum is achieved for $T=Z$. 

For $\pi\ll\prior$ such that $\mathcal{D}_f(\pi,\prior)<\infty$, $Z= \frac{\mathrm{d}\pi}{\mathrm{d}\prior}(\omega)$ for $\omega\sim \prior$ is $\prior$ integrable. Hence Lemma 14.2 implies 
\begin{align*}
\mathcal{D}_f\left(\pi, \prior\right)=&\sup_{T\neq 0}\left\{\pi \left[ f'(T) \right] +\prior\left[-Tf'(T) + (T-1) f'(\prior \bk{T}) +f(T) - f(\prior\bk{T})\right] \right\}\\
=&\sup_{T\neq 0}\left\{\pi \left[ f'(T) \right] -\left(\prior\left[
f^*\circ f'(T)\right] -f^*\circ f'\left(\prior\bk{T}\right) + f'(\prior\bk{T})\right)
\right\}
\end{align*}
where we used $xf'(x)- f(x) = f^*\circ f' (x)$ twice to obtain the second equality.
Consider the change of variable $D= f'\circ T$ which maps $\prior$-integrable random variables to the set of functions
\begin{equation*}
\mathcal{T}:=\left\{D \mid \forall h \in \text{Supp}\left(\pi\right), ~ f'(0)\leq D(h) \leq f'(\infty), ~ \prior\bk{\lvert{f^*}'\circ D \rvert}<\infty \right\}.
\end{equation*}
Note that this change of variable is well defined, since $f$ being twice differentiable implies that $f'$ has inverse ${f^*}'$. The bound becomes
\begin{align*}
\mathcal{D}_f\left(\pi, \prior\right)=\sup_{D\in \mathcal{T}} \pi \left[ D \right] - \pth{\prior\bk{f^*\circ D} - \tilde{\Delta}_f(D)}.
\end{align*}

We extend outside of $\mathcal{T}$ by checking the behaviour of the bound when some conditions are broken. We can first relax the condition that $D\geq f'(0)$ since ${f^*}'\circ D = {f^*}'(\max(f'(0),D))$ and $f^*\circ D = f^*(\max(f'(0), D))$. Then, we can relax the hypothesis that ${f^*}'\circ D$ is $\prior$ integrable. If it is not the case, then $\tilde{\Delta}_f(D) = 0$ and the Fenchel--Young inequality from \Cref{eq:ch2:loose_bound} implies that $\mathcal{D}_f(\pi, \prior) \geq \pi[D]$. Hence no element higher than $\mathcal{D}_f(\pi,\prior)$ is added in the sup, and hence this does not increase the sup. Thus
\begin{align*}
\mathcal{D}_f(\pi,\prior)&=\sup_{D\in \mathcal{T}}\left\{\pi\bk{D} -\pth{\prior\bk{f^*\circ D} - \tilde{\Delta}_f(D)}\right\}\\
&= \sup_{\substack{D<f'(\infty),\\\prior\bk{D}<\infty}}\left\{\pi\bk{D} -\pth{\prior\bk{f^*\circ D} - \tilde{\Delta}_f(D)}\right\}\\
&= \sup_D\left\{\pi \bk{D} -\pth{\prior\bk{f^*\circ D} - \tilde{\Delta}_f(D)}\right\}.
\end{align*}
This implies Fenchel--Young's inequality, which implies the result.
\end{proof}

\begin{remark}
\label{rmk:ch2:tight_improves_on_loose}
\Cref{eq:ch2:tight_upper_bound} gives a better approximation than equation \Cref{eq:ch2:loose_bound}. Indeed, by definition,
\begin{align*}
f^*(t)= \sup_{x>0}xt - f(x) \geq t
\end{align*}
since $f(1) = 0$. Hence $\forall t$, $t - f^*(t)\leq 0$. This implies that $\forall D$, $\tilde{\Delta}_f\geq 0$.
\end{remark}

\begin{remark}
\label{rmk:ch2:tight_minimised_is_legendre}
Consider a convex function $f$ and $D$ satisfying both the assumptions of \Cref{thm:ch2_1:theorem_optim_c} and \Cref{thm:ch2_1:tighter_approx}. Then the inequality in \Cref{eq:ch2:tight_upper_bound} is an equality. Notably, it is met for any $c^*$ satisfying the condition given in \Cref{thm:ch2_1:theorem_optim_c}.
\end{remark}

\begin{proof}
This is a consequence of the fact for a given $c$, the right hand side of \Cref{eq:ch2:tight_upper_bound} is upper bounded by the right hand side of \Cref{eq:ch2:loose_bound}, and lower bounded by $\overline{D}_f^*$. Since the minima of the right hand side of \Cref{eq:ch2:loose_bound} matches $\overline{D}_f^*$, so must the minima of the right hand side of \Cref{eq:ch2:tight_upper_bound}.
\end{proof}

If one considers Kullback--Leibler divergence, the evaluation of the right hand side of \Cref{eq:ch2:tight_upper_bound} gives $-\log(\prior\bk{\exp(-D)})$ for any $c$, exactly matching the exact Legendre transform. This shows that the refined bound can significantly improve on the standard bound.

\begin{remark}
\label{rmk:ch2:tight_only_for_more_than_second_moment}
The condition $1/f''$ concave implies that $$\lim\inf f^*(t) /t^2>0.$$
This can be interpreted as a requirement that $D$ has at least second order moment for \Cref{thm:ch2_1:tighter_approx} to yield non vacuous bounds.
\end{remark}

\begin{proof}
Since $f$ is convex and twice differentiable, it follows that $f''>=0$. Therefore, the concave function $1/f''$ is concave and positive on $\R_+^*$.

Let us show that this implies that $1/f''$ is non decreasing.
Suppose that there exists $x_1> x_2 > 0$, $1/f''(x_1) < 1/f''(x_2)$, the concavity of $1/f''$ implies that for all $x> x_1$, 
\begin{align*}
\frac{1}{f''(x)} \leq -\pth{\frac{1}{f''(x_2)} - \frac{1}{f''(x_1)}}\frac{x-x_2}{x_1 - x_2} + \frac{1}{f''(x_2)}.
\end{align*}
As the right hand side goes to $-\infty$ as $x\rightarrow \infty$, this is impossible, and hence $1/f''$ must be non decreasing.

There thus exists $t_0>0$, $\alpha = 1/f''(t_0)>0$ such that for all $t>t_0$, $1/f''(t)\geq\alpha$. Hence $f''(t)\leq\alpha^{-1}$, which implies $f'(t) \leq \alpha^{-1}(t-t_0) + f'(t_0)$, and hence, using the fact that ${f^*}'$ is increasing and the inverse of $f'$, that $t \leq {f^*}'\pth{\alpha^{-1}(t-t_0) + f'(t_0)}$. Thus for all $t > f'(t_0)$, we have $\alpha t + t_0 - \alpha f'(t_0) \leq {f^*}'(t)$. By integration, it follows that $f^*(t)/t^2 \geq \alpha / 2 + O(1/t)$ for all $t> f'(t_0)$. Taking $t\rightarrow \infty$ concludes the proof.
\end{proof}

\subsection{A temperature degree of freedom}
\label{sec:ch2:temperature_deg_freedom}
Another way to improve the resulting bound is the introduction of a scale degree of freedom. Due to its close relationship to the Gibbs temperature in the case where the $f$-divergence is the KL divergence, we call this degree of freedom the temperature and note it $\lambda$. This degree of freedom can be introduced in two equivalent ways; either by replacing the convex function $f$ by $\lambda f$, or by replacing the generalised generalisation gap $D$ by $\lambda^{-1} D$. In both cases, this result in an extended form of the bound as
\begin{equation*}
\prior\bk{D} \leq \lambda \prior\bk{f^*\circ(\lambda^{-1}D)}+ \lambda\mathcal{D}_f(\pi, \prior).
\end{equation*}

We now give the most general form of Fenchel--Young's inequality with $f$-divergence penalisation in the following theorem.\smallskip

\begin{theorem}
\label{thm:ch2_1:pre_Catoni_bounds}
For $\prior$ a probability measure on $\mathcal{H}$, for $f$ a convex function such that $f(1)=0$, then for any lower bounded, measurable function $D$, for all $\lambda >0$, for all $c\in\R$,

\begin{equation}
\label{eq:ch2_1:loose_2_degrees}
\pi\bk{D} \leq \lambda \prior\bk{f^*(\lambda^{-1}(D-c))}+c+\lambda \mathcal{D}_f(\pi, \prior).
\end{equation} 

Moreover, if $f$ is twice differentiable such that $1/f''$ is concave, then 
\begin{equation}
\label{eq:ch2_1:tight_2_degrees}
\pi\bk{D} \leq \lambda \prior\bk{f^*(\lambda^{-1} (D-c))} - \tilde{\Delta}(\lambda^{-1}(D-c)) + c+ \lambda \mathcal{D}_f(\pi, \prior).
\end{equation}
\end{theorem}

\begin{remark}
\label{rmk:ch2:moment_assumption}
To use \Cref{thm:ch2_1:pre_Catoni_bounds} to bound $\pi[D]$ simultaneously for all $\pi$, then there must exist $c$ such that the $f^*$ moment of $\lambda^{-1}(D-c)$ is upper bounded. This implies that in the case of KL, the exponential moment assumption can not be weakened, since $f^*(t)= \exp(t-1)$.
\end{remark}

\begin{figure}[htbp]
\centering
\includegraphics[width=0.98\linewidth]{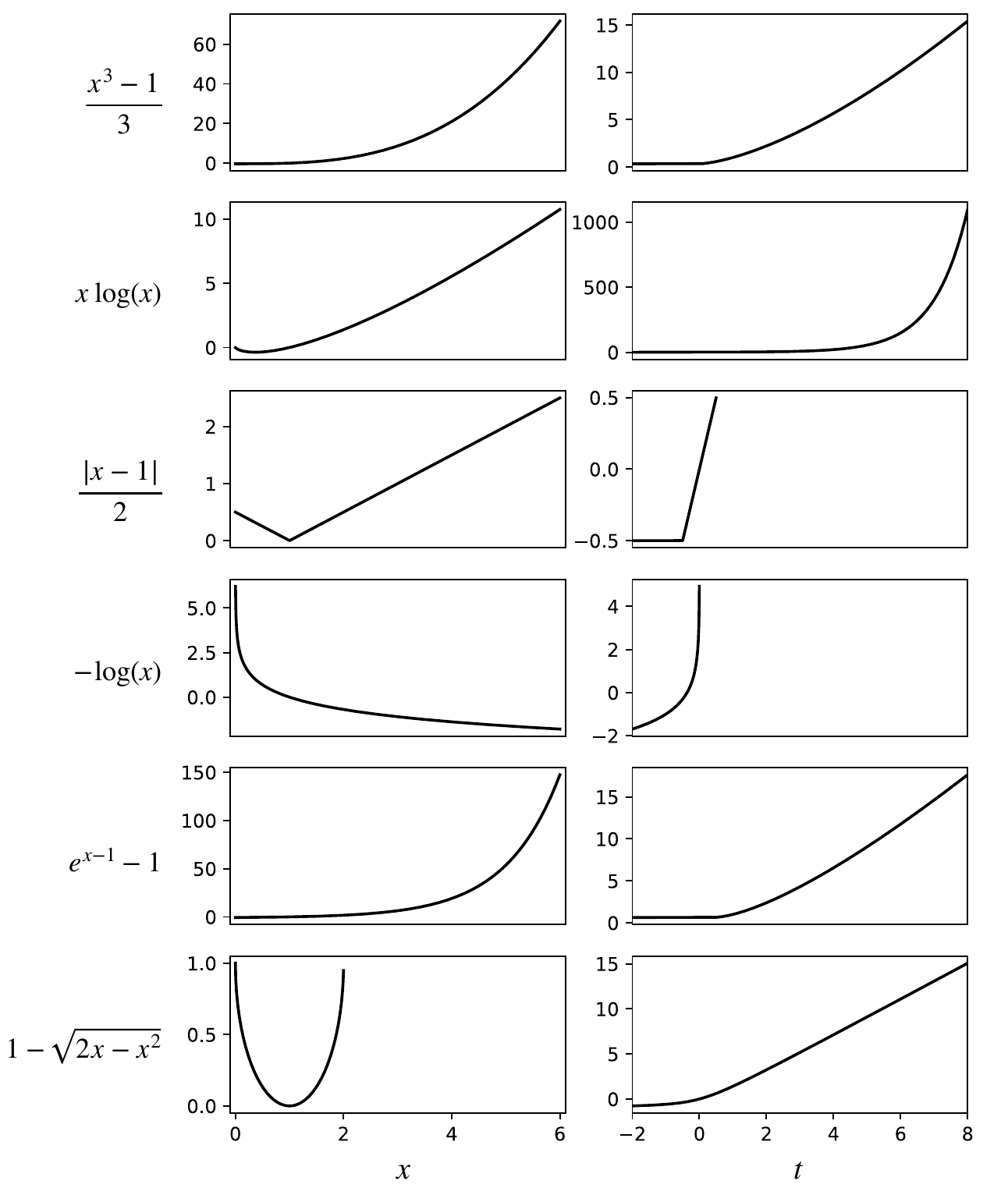}
\caption{Various convex functions \ensuremath{f} satisfying \ensuremath{f(1)=0} (on the left column) and their corresponding Legendre transform \ensuremath{f^*} (on the right column). The first row corresponds to a power 3 divergence, resulting in \ensuremath{f^*} behaving asymptotically as \ensuremath{t^{3/2}}. The second row corresponds to Kullback--Leibler divergence, and result in an exponential Legendre transform. The third and fourth rows describe weak penalisation (\ensuremath{f'(\infty) < \infty}), leading to \ensuremath{f^*} with upper bounded support (note that the third row corresponds to the Total variation distance). The fifth and sixth rows describe strong penalisation, resulting in slowly increasing \ensuremath{f^*}.}
\label{figure:ch2_1:various_f_and_f_div}
\end{figure}

\Cref{thm:ch2_1:pre_Catoni_bounds} states that we can control the average of the generalised generalisation gap $D$ over all probability measures $\pi$ from two terms~: a measure of the distance between $\pi$ and $\prior$, and what is morally a moment of the random variable with respect to $\prior$. These two terms offer a trade-off between the type of penalisation considered - controlled by how fast $f$ grows - and the strength of the moment assumption - controlled by how fast $f^*$ grows (see \Cref{figure:ch2_1:various_f_and_f_div}). The more the $f$-divergence discriminates between $\pi$ and $\prior$, the weaker is the moment needed. On the other hand, if strong moment assumptions can be made on the random variable, one can control its mean over $\pi$ for a larger set of probability measures.

The bounds can be optimised on two degrees of freedom, the positional parameter $c$ and a scale parameter $\lambda$. \Cref{thm:ch2_1:theorem_optim_c} implies that the first optimisation factor can recover the optimal additive bound for regular $f$ and generalisation gap with bounded ${f^*}'$ moments\footnote{We conjecture that this assumption can be relaxed, in the sense that if $\forall c$, $\prior\bk{{f^*}'(D-c)} = \infty$, then the Legendre transform is infinite. Whether there are settings where $\exists c_1$ such that $\prior\bk{{f^*}'(D-c_1)}<1$ but not $c_2$ such that $\prior\bk{{f^*}'(D-c_2)}>1$ and how the bound would behave in such cases is also an open question.}. Moreover, it states that optimisation on $c$ is related to the normalisation problem for the probability measure reaching the upper bound (the maximiser in the definition of the Legendre transform).

Although the bounds can be optimised on two degrees of freedom, it might not be possible to apply this double optimisation procedure. We do not have clear arguments to favour optimising with respect to $c$ over optimising with respect to $\lambda$ or vice-versa. We note however that most of the bounds we examined proved easier to optimise on the scale parameter rather than on the positional degree of freedom.\medskip

\begin{remark}
\label{rmk:ch2:inverse_f_div_is_f_div}
For all $f\in \mathcal{F}$, the reverse $f$-divergence $\pi_1, \pi_2 \to \mathcal{D}_f(\pi_2,\pi_1)$ is a $f$-divergence for $\tilde{f}(t)= t\times f\left(1/t\right)$. Therefore, \Cref{thm:ch2_1:pre_Catoni_bounds} provides change-of-measure inequalities for the reverse $f$-divergence.
\end{remark}

\begin{remark}
\label{rmk:ch2:bounded_f_differential}
Embedded in the bounds of \Cref{thm:ch2_1:pre_Catoni_bounds} is the condition that $\lambda (D-c)<f'(\infty)$ $\prior$-almost surely (see \Cref{remark:ch2_1:f_star_properties}). As a consequence, whenever $f'(+\infty)\neq +\infty$, the generalisation gap $D$ needs to be upper bounded for the bounds to be useable. Whenever this is the case, we find it good practice to choose $f$ such that $f'(\infty)=0$ and as such, we can apply the bound to $\lambda(D-D_{\max}-c)$ for $\lambda>0, c>0$.
\end{remark}

\begin{remark}
\label{rmk:ch2:minimisation_is_a_legendre_transform}
One can reinterpret the minimisation on $\lambda$ and $c$ for every bound of the form \Cref{eq:ch2_1:loose_2_degrees} in term of Legendre transforms. Indeed, for $\lambda >0$ and $c\in\R$, consider
\begin{align*}
L_c:\lambda \to \lambda \prior\bk{f^*\circ(\lambda^{-1} (D-c))},\\
L_\lambda : c\to \lambda \prior\bk{f^*\circ(\lambda^{-1} (D-c))}.
\end{align*}
Then both $L_c$ and $L_\lambda$ are convex functions, and the minimisation of the bound on $\lambda$ yields 
\begin{align*}
\pi[D] \leq -L_c^*(- \mathcal{D}_f(\pi, \prior)) + c,
\end{align*}
while the minimisation of the bound on $c$ yields
\begin{align*}
\pi[D] \leq -L_\lambda^*(-1) + \lambda \mathcal{D}_f(\pi, \prior).
\end{align*}
Moreover, if
\begin{align*}
L: \lambda, c\to \lambda \prior\bk{f^*\circ(\lambda^{-1} (D-c))}
\end{align*}
is convex, then the bound can be interpreted as
\begin{align*}
\pi[D] \leq -L^*\pth{\begin{pmatrix}- \mathcal{D}_f(\pi, \prior)\\-1\end{pmatrix}}.
\end{align*}
A similar argument can be used for the bound of form \Cref{eq:ch2_1:tight_2_degrees}, although in this case, the functions $L_c$ and $L_\lambda$ might not be convex any longer.
\end{remark}

\begin{remark}
\label{rmk:ch2:upperbound_f_star}
To define the Legendre transform of $f$ in \Cref{eq:ch2_1:legendre_f}, we consider a suprema on $x\in\R_+$. This is equivalent to extending $f$ to $\R$ by setting $f(x) = +\infty $ for all $x<0$ (any negative $x$ is thus ruled out since it leads to $-\infty$ in the bound). As noted in \Cref{remark:ch2_1:f_star_properties}, this introduces a threshold at $f'(0)$ in the values of $f^*$; that is to say, $f^*(x) = -\lim\inf_{x\rightarrow 0} f(x):= f(0)$ for all $x \leq f'(0)$. Hence the functional $D$ can be replaced by $\max(D, f'(0))$.

This threshold can be problematic when trying to optimise the bounds on the two degrees of freedom. A way to obtain more tractable bounds is to consider other convex extensions of $f$ to $\R$ in the definition of $f^*$. If $f'(0) > -\infty$, $f$ can be extended for $x< 0$ by $\tilde{f}(x) = f(0) + x f'(0)$. If $f''(0) < \infty$ moreover, $\tilde{f}(x) = f(0) + xf'(0) + \frac{x^2}{2}f''(0)$ for all $x<0$ also provides a convex extension of $f$. Specific $f$ might also have natural extensions (\emph{e.g.} power functions). Since Fenchel--Young's inequality remains valid for these $\tilde{f}^*$, the upper bounds of form \Cref{eq:ch2_1:loose_2_degrees} also remain valid. While these will result in looser bounds, the added tractability might result in better bound after optimisation.
\end{remark}

\section{Application to Learning Theory}\label{sec:ch2_1:fdiv}
\subsection{Some more PAC-Bayesian bounds}
We now explore how \Cref{thm:ch2_1:pre_Catoni_bounds} can be leveraged in learning theory. So far, the change-of-measure was performed for any function $D$. To obtain PAC-Bayes bounds, one can replace $D$ by the generalisation gap $D = \trisk - \risk$ (or $D = \max(\trisk - \risk, 0)$ if the risk is not bounded) in \Cref{eq:ch2_1:loose_2_degrees} to obtain, $\forall c\in\R$, $\forall \lambda \in\R_+$, $\forall\prior$, with probability higher than $1- \delta$, $\forall \pi\ll\prior$
\begin{equation*}
\pi\bk{ \trisk} \leq \pi\bk{\risk} + \frac{ \lambda \mathbb{P}\bk{\prior\left[f^*\circ (\lambda^{-1}(D-c)) \right]}}{\delta} +\lambda\mathcal{D}_f(\pi,\prior) +c.
\end{equation*}
This is simply a consequence of Markov's inequality on the moment term. Note that in this expression, the degree of freedom $\lambda$ and $c$ must be set before using Markov's inequality on the term of the right hand side independent of $\pi$. While the optimisation on $\lambda$ depends on the value of $\pi$ and hence couples the two terms of the Fenchel--Young inequality, the optimal choice on $c$ is independent on $\pi$ and can therefore be put into the expected value. This yields the improved PAC-Bayes bound, stating that $\forall \lambda \in\R_+$, $\forall\prior$, with probability higher than $1- \delta$, $\forall \pi\ll\prior$
\begin{equation*}
\pi\bk{ \trisk} \leq \pi\bk{\risk} + \frac{ \mathbb{P}\bk{\inf_{c}\lambda\prior\bk{f^*\circ (\lambda^{-1}(D-c))} + c}}{\delta} +\lambda\mathcal{D}_f(\pi,\prior).
\end{equation*}
Note that concentration inequalities other than Markov can be used to improve the resulting PAC-Bayes bound. For instance, for $\mathcal{D}_f = \KL$, Catoni's bound
\begin{equation}
\label{eq:ch1:catoni}
\PB_\lambda(\pi, \risk, \prior, \delta) := \pi\bk{\risk} + \lambda\KL(\pi, \prior) -\lambda\log(\delta)+\frac{1}{8n\lambda}
\end{equation}
can be recovered using a Chernoff bound. When dependent data is considered, adapted concentration inequalities should be used to bound the quantiles of $$\inf_{c}\prior\bk{\lambda f^*\circ (\lambda^{-1}(D-c))} + c.$$

The general form considered in \Cref{thm:ch2_1:pre_Catoni_bounds} can also be leveraged to obtain tighter PAC-Bayes bound using the generalised generalisation gap approach developed by \cite{begin2016pac}. Considering generalisation gaps of form $D(\omega) = \Delta(\trisk(\omega), \risk(\omega))$ with $\Delta$ a convex function, one can "inverse" the $\Delta$ function through $\Delta^{-1}(t,y) = \sup\curl{x\mid \Delta(x,y)\leq t}$. It then follows from Jensen's inequality and \Cref{thm:ch2_1:pre_Catoni_bounds} that $\forall \lambda>0$, $\forall \prior$, $\forall \delta$, with probability at least $1-\delta$, $\forall \pi\ll\prior$, 
\begin{equation}
\label{eq:ch2:PAC_Bayes_f_div_Delta}
 \pi\bk{\trisk} \leq \Delta^{-1}\pth{\frac{\mathbb{P}\bk{\inf_c
 \prior\bk{f^*\pth{\lambda^{-1}\Delta\pth{\trisk, \risk} -c}} + c}}{\delta} +\lambda \mathcal{D}_f(\pi, \prior), \pi\bk{\risk}}
\end{equation}
where once again, Markov's inequality can be replaced with a more strategic concentration inequality, and a fixed $c$ might be used for convenience.\medskip

\begin{remark}
\label{rmk:ch2:optimise_on_Delta_fun}
The change-of-measure inequalities of \Cref{thm:ch2_1:pre_Catoni_bounds} are valid \emph{for all} generalised generalisation gaps $D$. As a consequence, it follows that the bound can \emph{theoretically} be optimised on the convex function $\Delta$, and even on all $f$-divergences, resulting in 
\begin{equation}
\label{eq:ch2_1:optimised_bound}
\forall \pi\ll\prior, \pi[\trisk] \leq \inf_{f\in\mathcal{F}}\inf_{\Delta} \Delta^{-1}\pth{\prior\bk{f^*\pth{\Delta\pth{\trisk, \risk}}} + \mathcal{D}_f(\prior, \pi), \pi[\risk]}
\end{equation}
where $\mathcal{F}$ is the set of all convex functions of $\R_+$ to $\R$ such that $f(1)=0$, and the minima on $\Delta$ is taken on all lower bounded convex functions. Note that this formulation recovers both degrees of freedom, and should be quite tight. 

However, to obtain a PAC-Bayes bound, it is necessary to upper bound the quantiles of the right hand side \emph{simultaneously} for all posterior distributions $\pi$. This, in the general case, prevents optimisation on a degree of freedom whenever the optimal value depends on the posterior distribution (this couples the value of the bound to $\pi$) - and the analysis is blocked when there is no closed form expression for the minima (the coupling is unknown). Hence such optimised forms as \Cref{eq:ch2_1:optimised_bound} are in the general case of little use in obtaining PAC-Bayes bounds. For some tractable bounds, it is possible to upper bound the quantiles for all $\pi$ \emph{after} optimisation on the temperature degree of freedom $\lambda$ (see \Cref{sec:ch2:some_change_of_measure_inequalities}).
\end{remark}

\subsection{From moment assumption to penalisation}
We now consider a setting where the generalised generalisation gap $D$ is fixed. We study in this section the task of reverse engineering assumptions on the $M$-moment of the generalisation gap into PAC-Bayes bound. In plain words, the question we are trying to answer is whether we can transform an assumption of form $\forall h \in\mathcal{H}$, $\mathbb{P}\bk{M\circ D(h)} \leq \alpha$ into a PAC-Bayes bound. We show that such a strategy is indeed possible if $M$ goes faster to infinity than linearly, and study how the form of $M$ impacts the bound.

Let us assume that $D>0$, and that the function $M$ satisfies $\lim_{t\rightarrow+\infty} M(t)/t=+\infty$ and $M(t)>t$. Denote $M_{-}$ the lower convex envelope of $M$. Note that $M_{-}$ also satisfies $\lim_{t\rightarrow + \infty} M(t)/t = +\infty$ (since the conditions imply that for all $a >0$, $\exists b$ and $t_a$ such that $M(t) > at +b$ $\forall t> t_a$). Since for all values of $D$, $0 \leq M_{-}\leq M$, it follows that $\mathbb{P}\bk{M_{-}\circ D}\leq \alpha$. Using $f= M_{-}^* - M_{-}^*(1)$ in \Cref{thm:ch2_1:pre_Catoni_bounds} implies that 
\begin{equation*}
\pi[D] \leq \prior\bk{M_{-} \circ D + M_{-}^*(1) \times D} + \mathcal{D}_{M_{-}^* - M_{-}^*(1)}.
\end{equation*}
Using the fact that $M_{-}(t)> t$ (inherited from the condition on $M$), this implies that $\forall\pi$
\begin{equation*}
\pi[D] \leq (1+ M_{-}^*(1))\prior\bk{M_{-} \circ D } + \mathcal{D}_{M_{-}^* - M_{-}^*(1)}(\pi,\prior).
\end{equation*}
Finally, using Fubini in conjunction with Markov's inequality implies that with probability at least $1-\delta$, for all $\pi$,
\begin{equation}
\label{eq:ch2:PAC_Bayes_for_any_assumption}
\pi\bk{D}\leq \frac{(1+M_{-}^*(1))\alpha}{\delta} + \mathcal{D}_{M_{-}^* - M_{-}^*(1)}(\pi,\prior).
\end{equation}

If $D$ is of form $\Delta(\trisk, \risk)$, the same Jensen argument as in \Cref{eq:ch2:PAC_Bayes_f_div_Delta} can be used, leading to

\begin{equation}
\label{eq:ch2:PAC_Bayes_for_any_assumption_Delta}
\pi\bk{\trisk}\leq \Delta^{-1}\pth{\frac{(1+M_{-}^*(1))\alpha}{\delta} + \mathcal{D}_{M_{-}^* - M_{-}^*(1)}(\pi,\prior), \pi\bk{\risk}}
\end{equation}
holding simultaneously for all $\pi$ with probability at least $1- \delta$.

\section{Some change-of-measure inequalities}
\label{sec:ch2:some_change_of_measure_inequalities}
\subsection{Standard \ensuremath{f}-divergence}
\label{sec:ch2:change_of_measures_std_f_div}

We apply \Cref{thm:ch2_1:theorem_optim_c} to the most popular $f$-divergences found in the literature. \Cref{table:ch2_1:fdiv_bounds} presents a summary of all the resulting change-of-measure inequalities. Note that the optimal values of $\lambda$ and the Legendre transform $f^*$ are gathered, when available, in \Cref{table:app_2:proof_fdiv}.

\renewcommand{\cellalign}{cl}

\begin{table}[htbp]
\centering
\small
\caption{Bounds for typical \ensuremath{f}-divergences. We denote \ensuremath{D_+:=\max(D,0)}.
For power-divergences, \ensuremath{q} is such that \ensuremath{\frac{1}{q}+\frac{1}{p} =1}.
For Lin's measure, \ensuremath{f_\vartheta} is given by \ensuremath{f_\vartheta(t)= (\vartheta t\log(t\vartheta)- (\vartheta t + 1- \vartheta)\log(\vartheta t + 1 - \vartheta)-\vartheta\log(\vartheta)}. }
\vspace{1em}
\renewcommand{\arraystretch}{2.2}
\begin{centering}
\begin{tabular}{|p{1.9cm}|p{1.4cm}|p{8.3cm}|p{1.1cm}|}\hline\hline
$f$-div
& $f(t)=$
& $\pi\bk{D}\leq \dots$
& $c, \lambda$\\\hline\hline
KL
& $t\log(t)$
& \( \lambda\log\prior\bk{\exp\left(\lambda^{-1} D\right)}+\lambda\KL(\pi, \prior)\)
& $\lambda>0$
\\
\hline
\makecell{Power-$p$,\\ $1<p\leq 2$}
& \( t^p-1\)
& $
 \prior \bk{D_+^{q-1}}^{p-1}
+\left(\prior \bk{D_+^q} - \prior\bk{D_+^{\frac{q}{p}}}^p\right)^\frac{1}{q}
\mathcal{D}_{f_p}\left(\pi, \prior\right)^\frac{1}{p}
$
& 
\\
\hline
\makecell{Power-$p$,\\$1<p$}
& $\displaystyle t^p-1$
& $
\displaystyle c+\prior \bk{(D-c)_+^q}^{\frac{1}{q}}\left(1+\mathcal{D}_{f_p}\left(\pi, \prior\right)\right)^{\frac{1}{p}}
$
& $c\in \R$
\\
\hline
Pearson $\chi^2$
& \(t^2-1\)
&\(\displaystyle \prior \bk{D_+} +
\variance_{\prior}\bk{D_+}^{\frac{1}{2}}
\chi^2\left(\pi, \prior\right)^\frac{1}{2}\)
&
\\
\hline
\makecell{Power-$p$,\\$0<p<1$}
& \(1- t^p\)
& \(D_{\max}+c -\prior\bk{\left(D_{\max}-D+c\right)^q}^{\frac{1}{q}} \left(1- \mathcal{D}_{f_p}(\pi,\prior)\right)^{\frac{1}{p}}\)
& $c>0$\\
\hline
\makecell{Power-$p$,\\ $p<0$}
& \(t^p-1\)
& \(D_{\max}+c - \prior\bk{\left(D_{\max}-D+c\right)^q}^{\frac{1}{q}} \left(1+ \mathcal{D}_{f_p}(\pi,\prior)\right)^{\frac{1}{p}}\)
& $c\geq 0$
\\
\hline
TV
& $\abs{t-1}/2$
& \(\displaystyle D_{\max}+\prior\bk{\max\pth{D-D_{\max}, -\lambda}} +\lambda \textup{TV}(\pi,\prior)\)
& $\lambda>0$
\\
\hline
\makecell{Squared\\Hellinger}
& \(1- \sqrt{t}\)
& \(\displaystyle D_{\max}+c-\left(1-H^2(\pi,\prior)\right)^2 \prior\bk{ \frac{1}{D_{\max}-D+c}}^{-1}
\)
& $c>0$
\\
\hline
\makecell{Reverse\\Pearson}
& $t^{-1}-1$
& \(D_{\max}+c- \frac{\prior\bk{\sqrt{c+D_{\max}-D}}^2}{1+\chi^2(\prior,\pi)}\)
& $c>0$
\\
\hline
\makecell{Reverse\\KL}
& \(-\log(t)\)
& \(D_{\max}+c- \exp\pth{\prior\bk{\log\pth{D_{\max}-D+c}}}- \KL\pth{\prior, \pi)}\)
& $c>0$
\\
\hline
\makecell{Lin's\\measure\\($\vartheta\in]0,1[$)}
& $f_\vartheta(t)$
& $\begin{aligned}
&D_{\max}+c\\
&-\lambda (1-\vartheta)\prior\bk{\log\left(1- \exp\left(\lambda^{-1}\vartheta^{-1}(D-D_{\max}-c)\right)\right)}\\
&+\lambda\pth{L_\vartheta(\pi,\prior)+(1-\vartheta)\log(1-\vartheta)- \vartheta\log(\vartheta)}
\end{aligned}$
& \makecell{$\lambda>0$,\\ $c>0$}
\\
\hline
\makecell{Jensen-\\Shannon}
& $f_{\vartheta=\frac{1}{2}}(t)$
& \makecell{$D_{\max} + c- \lambda\prior\bk{ \frac{1}{2}\log\left(1- \exp\left(2 \lambda^{-1}(D-D_{\max}-c)\right)\right)}$\\$ +\lambda\text{JS}(\pi,\prior)$}
& \makecell{$\lambda>0$,\\ $c>0$}
\\
\hline
\makecell{Vincze-\\Le Cam}
&$ \frac{2-2t}{t+1}$
& $ 2~(D_{\max}+c) + \prior\bk{-D} - \frac{4\prior\bk{\sqrt{c+D_{\max}-D}}^2}{2+\text{VC}(\pi,\prior)}$
& $c>0$
\\\hline
&$e^{t-1}-1$
% &$$\begin{aligned}
% &\prior\bk{(D - c)_+ + \lambda/e)\log(\lambda^{-1} ((D - c)_+ + \lambda/e))} + \lambda \\
% &+ c- \lambda/e+ \lambda\mathcal{D}_f(\pi, \prior)
% \end{aligned}$$
&$\begin{aligned}
&\prior\bk{(D - c)_+ + \lambda)\log( ((D - c)_+ + \lambda))}\\
&- (1+\log(\lambda))\prior\bk{(D - c)_+ + \lambda)}\\
&+ c+ (e-1)\lambda+ e\lambda\mathcal{D}_f(\pi, \prior)
\end{aligned}$
&$\begin{aligned}&c\in\R\\& \lambda > 0\end{aligned}$
\\
\hline
\end{tabular}
\end{centering}
\label{table:ch2_1:fdiv_bounds}
\end{table}

\begin{proof}
The proof for each bound follows the same pattern: for each $f$-divergence, compute $f^*$, check whether $(1/f'')$ is concave, then apply accordingly either \cref{eq:ch2_1:tight_2_degrees} or \cref{eq:ch2_1:loose_2_degrees} to obtain:

\begin{equation*}\pi\bk{D} \leq c+ \lambda\text{B}(\lambda^{-1}(D-c)) +\lambda\mathcal{D}_f(\pi,\prior).
\end{equation*}
Then optimise on $\lambda$ and $c$ whenever feasible. We therefore sum up the proofs in \Cref{table:app_2:proof_fdiv} which details the form of the $f^*$ as well as the optimal value of $\lambda$ when computable.

\begin{table}[phtb]
\centering
\normalsize
\renewcommand{\arraystretch}{2.2}
\caption{$f^*$, $\lambda^*$  optimising change of measure inequalities.}
% \vspace{1em}
\begin{tabular}{|l|c|c|}\hline
$f$-div
& $f^*(t)=\dots$
& $\lambda^*$
\\\hline\hline
KL
& $\exp\left(t-1\right)$
& 
\\
\hline
Power-$p$, $1<p\leq 2$
& $\frac{p^{1-q}}{q}\max(t,0)^q+1 $
& $p\left(\frac{\mathcal{D}_{f_p}(\pi,\prior)}{\prior\bk{D_+^q} - \pi\bk{D_+^{\frac{q}{p}}}^{p}}\right)^{\frac{1}{q}} $
\\
\hline
Power-$p$, $2<p$
& $ \frac{p^{1-q}}{q}\max(t,0)^q+1 $
& $p\left(\frac{1+\mathcal{D}_{f_p}(\pi,\prior)}{\prior \bk{h^q}}\right)^{\frac{1}{q}} $
\\
\hline
Pearson $\chi^2$
& \(\sqrt{2}\max(t,0)^2+1\)
& $2\left(\frac{\mathcal{D}_{f_2}(\pi,\prior)}{\textbf{Var}_\pi\bk{D_+}}\right)^{\frac{1}{2}} $
\\
\hline
Power-$p$, $0<p<1$
& $ \begin{cases}-1+\frac{p^{1-q}}{-q}(-t)^q & t<0\\+\infty & \text{else}\end{cases} $
& $p \left(\frac{1- \mathcal{D}_{f_p}(\pi,\prior)}{\prior\bk{(-D)^q}}\right)^{\frac{1}{q}}$
\\
\hline
Squared Hellinger
& $\begin{cases}
-1 +\frac{1}{4}(-t)^{-1} & t<0\\
+\infty & t\geq 0
\end{cases}$
& $\frac{\pi\bk{(-D)^{-1}}}{2-2H^2(\pi,\prior)}$
\\
\hline
Power-$p$, $p<0$
& $\begin{cases}1-\frac{(-p)^{1-q}}{q}(-t)^q & t\leq 0\\+\infty & \text{else}\end{cases}$
& $(-p)\left(\frac{1+ \mathcal{D}_{f_p}(\pi,\prior)}{\prior\bk{(-D)^q} }\right)^{\frac{1}{q}}$
\\
\hline
Reverse Pearson
& $\begin{cases}
1- 2(-t)^{1/2} & t\leq 0\\
+\infty & \text{else}
\end{cases}$
& $\left(\frac{1+ \chi^2(\prior, \pi)}{\pi\bk{\sqrt{(-D)}}}\right)^{2}$
\\
\hline
Total Variation
& $\begin{cases}
-\frac{1}{2} & t <-\frac{1}{2}\\
t & -\frac{1}{2}\leq t\leq \frac{1}{2}\\
+\infty & t>\frac{1}{2}
\end{cases}$
&
\\
\hline
Reverse KL
& $\begin{cases}-(1+\log(-t)& t<0\\
+\infty &\text{else}
\end{cases}$
& $\exp\left(\text{KL}(\pi,\nu) - \pi\bk{\log(-D)}\right)$
\\
\hline
\makecell{Lin's measure,\\$0<\theta<1$}
& $(1-\theta) \log\left(\frac{1-\theta}{1- \theta\exp\left(t\theta^{-1}\right)}\right)$
&
\\
\hline
Jensen--Shannon
&
$-\frac{1}{2}\log\left(2-e^{2t}\right)$
&
\\
\hline
Vincze--Le Cam
& $\begin{cases}
-2 &t\leq -4\\
-4\sqrt{-t}-t+2& -4\leq t\leq 0\\
+\infty &\text{else}
\end{cases}$
& $\left(\frac{1+\frac{1}{2}\text{VC}(\pi,\prior)}{\prior[\sqrt{-D}]}\right)^2$
\\\hline
$e^{t-1}-1$ 
& $\begin{cases}
-1/e + 1&t\leq \frac{1}{e} \\
t\log(t) + 1 & \geq \frac{1}{e}
\end{cases}$
&
\\\hline
\end{tabular}
\label{table:app_2:proof_fdiv}
\end{table}

The Kullback--Leibler, power divergence for $1<p\leq 2$ and Pearson $\chi^2$ satisfy $1/f''$ concave, and we therefore use \Cref{eq:ch2:tight_upper_bound}. All the other bounds use \Cref{eq:ch2:loose_bound}.
For the total variation, it is simple to see that to get non trivial bounds, we need to pick $c\leq \frac{1}{2}-D_{\max}$. Diminishing $c$ to $c-\delta c$ decreases the integral by at most $\delta c$ (the threshold can only dampen the decrease), while the other term increases by $\delta c$. This implies that $c^*=\frac{1}{2}-D_{\max}$.

For the Vincze--Le Cam divergence, we are in the situation described in \Cref{rmk:ch2:upperbound_f_star}. The upper bound obtained through $\tilde{f}$ is much more tractable than the one obtained through $f$, and in particular, it can be optimised on the scale parameter $\lambda$. It is this bound through $\tilde{f}$ which we use to obtain the final bound. Using the Legendre transform of $f_{\lvert \mathbb{R}_+}$ yields this tighter, though less tractable, inequality for all $c\geq 0$, $\lambda>0$
\begin{align*}
\pi\bk{D} \leq  c &+ D_{\max} +  \prior\bk{ \mathbbm{1} \bk{ 4 \lambda - c \geq D^{-} } \left(4\lambda-4\sqrt{\lambda (c+D^{-})}+ (c+ D^{-}) \right)}
\\&+\lambda\left(\text{VC}(\pi,\prior) - 2\right).
\end{align*}
where $D^- = D_{\max}- D$.

\end{proof}

\Cref{eq:ch2:tight_upper_bound} recovers the exact Legendre transform of the KL divergence. The bound is also quite tight for Pearson's $\chi^2$- divergence. Indeed, \Cref{thm:ch2_1:theorem_optim_c} implies that for $D$ such that $D>0$ and $\prior\bk{D}\leq 1$, $\overline{\mathcal{D}}_{f}^*(D)$ has closed form expression $\frac{1}{2}\mathbb{V}_{\prior}\bk{D} + \prior\bk{D}$.

\begin{theorem}[Legendre transform of Pearson $\chi^2$ divergence]
\label{them:app2:pearson_chi2}
For a generalised generalisation gap $D$ satisfying $\prior[D^2]<\infty$, the Legendre transform of Pearson $\chi^2$ divergence can be obtained from \Cref{eq:ch2_1:minimisation_on_c}, \emph{i.e.}
\begin{equation}
\overline{\chi^2}^*(D) = \frac{1}{2}{\prior}\bk{(D-c)_+^2 +1} + c.
\end{equation}
Moreover, if $D$ satisfies $D>0$, $\prior[D] \leq 1$, the Legendre transform of Pearson $\chi^2$ divergence is
\begin{equation}
\overline{\chi^2}^*(D) = \frac{1}{2}\mathbb{V}_{\prior}\bk{D} + \prior[D].
\end{equation}
\end{theorem}
\begin{proof}
Let us prove the first statement of the theorem. The function $f^*(x) = \frac{x_+^2 + 1}{2}$ is differentiable, with derivative ${f^*}'(x) = x_+$. Since $\prior\bk{\abs{D}}<\infty$, this implies that $\prior\bk{{f^*}'(D)}< \infty$ and, considering the form of ${f^*}'$, that $1\leq\prior\bk{{f^*}'(D+1)}<\infty$. Hence the second part of \Cref{thm:ch2_1:theorem_optim_c} holds.

Then if $D>0$ and $\prior\bk{D} \leq 1$, it follows that $c^* = \prior\bk{D} -1 \leq 0$ is such that $D- c^*>0$ and hence $\prior\bk{{f^*}'(D-c^*)} = 1$. Hence $c^*$ minimises the bound. Evaluating the bound for $c^*$ finishes the proof.
\end{proof}

The bounds presented in \Cref{table:ch2_1:fdiv_bounds} are coherent with those obtained independently by \cite{ohnishi2021}. The last three are, to the best of our knowledge, the first change-of-measure inequalities obtained for these $f$-divergences.

For KL, one recovers the change-of-measure inequality established by \cite{csiszar1975divergence} and \cite{donsker1975large}. That bound is the starting point of the proof of the general PAC bound established by \cite{begin2016pac}, which recovers bounds obtained in \cite{langford2001bounds}, \cite{mcallester2003pac}, \cite{catoni2007pac} and \cite{alquier2016a}. %\citealp{dalalyan2012} provide some preliminary heuristics on the selection of $\lambda$; we will quickly discuss this point in \Cref{sec:ch2_1:lambda_selection}.

For power $p$ divergences with $p>1$, only moments of order $\frac{p}{p-1} = q$ for $D$ are needed rather than exponential moments, considerably lessening the assumptions needed on the loss $l$ and the underlying data distribution. When $1<p\leq 2$, the bounds we propose improve on those obtained in \cite{alquier2018simpler}. Indeed, these last bounds exactly match those we obtain through \Cref{eq:ch2_1:loose_2_degrees} for $c=0$ after minimisation on $\lambda$, which is looser than \Cref{eq:ch2_1:tight_2_degrees} which we consider. The bounds for $1<p\leq 2$ can be slightly simplified, noticing that $$\prior\bk{D_+^{q}} - \prior\bk{D_+^{\frac{q}{p}}}^p\leq \pi\left[D^{q}\right]- \prior\bk{D^{\frac{q}{p}}}^p.$$

For all the remaining $f$-divergences, $f'(\infty)<\infty$. Therefore the Legendre transforms of these $f$-divergences only take real values on bounded functions $D$. The bounds are of the form $D_{\max}$ minus a term involving the moment of $D_{\max} - D$.

For the power divergences with $0<p<1$, let us remark that when $\mathcal{D}_{f_p}(\pi,\prior)\rightarrow 0$, the bound is optimised for $c\rightarrow \infty$, while when $\mathcal{D}_{f_p}(\pi,\prior)\rightarrow 1$, the bound is optimised for $c\rightarrow 0$. A similar behaviour is observed for power divergences with $p<0$. It seems important to pick adequately $c\left(\mathcal{D}_{f_p}(\pi,\prior)\right)$ if one wishes to obtain tight bounds for all $\pi$.

For total variation, let us first remark that since the generator $f(x) = \frac{\abs{x-1}}{2}$ is not differentiable at $x=1$, it can not be approximated by a sequence of convex functions such that $1/f_n''$ is concave\footnote{Whenever $f'$ is not continuous at $x_0 > 0$, then $f''$ is a Dirac mass at $x_0$ and therefore $1 / f''(x_0)=0$. It follows that $1/f''$ has a local minima at $x_0$ since $f''\geq 0$, and therefore it can not be concave for any reasonable approximation.}. It is possible to minimise the bound on $c$, but we could not compute the optimal scale parameter.

Vincze-Le Cam's bound somewhat stands out as it involves $2 D_{\max}$ rather than $D_{\max}$. This is explained by the fact that the bound is not derived directly from \Cref{thm:ch2_1:pre_Catoni_bounds}, but results from \Cref{rmk:ch2:upperbound_f_star}, extending $f$ to $t\in(-1,0)$ by $f(t)=\frac{2-2t}{t+1}$.

\subsection{Change-of-measure with strong penalisation}
The strength of the penalisation considered in \Cref{eq:ch2:PAC_Bayes_for_any_assumption} depends on the strength of the moment assumption considered. Stronger penalisation will result in looser moment assumption, while on the other hand, strong moment assumption leads to weaker penalisation. The usual KL divergence being obtained for $f(x)= x\log(x)$ which grows slowly to infinity, it involves strong exponential moments on the generalisation gap. The trade-off between moment assumption and penalisation is also apparent for the power $f$-divergence, where $p$ power f-divergence results in the conjugate $q$ moment. Choices of $f$ such that $f(\infty) < +\infty$ (which implies that no choice of $f$ is super linear) leads to strict upper bounded generalisation gap requirement. On the other hand, choosing the fast growing $f(x) = e^{x-1} -1$ leads to the mild requirement of $x\log(x)$ bounds. Note that all moment requirements must be stronger than the first order moment, since $f^*(x) \geq x$ whenever $f(1) = 0$.

\cite{begin2016pac,alquier2018simpler} broke from the traditional bounded or bounded exponential moment requirement by obtaining bounds involving finite $p$-moments for all $p>1$. We go a step further by introducing two bounds involving strong penalisation and resulting in a $x\log(x)$-moment requirement or a first order moment requirement.

Our first bound considers an exponential $f$-divergence applied to positive generalisation gaps. For $D \geq 0$, $\forall \pi\ll\prior$,
\begin{equation}
\pi\bk{D} \leq \prior\bk{D\log(D)}- \prior\bk{D}\log\left(\prior\bk{ D} \right) +
\log\prior\bk{\exp\left(\frac{\mathrm{d}\pi}{\mathrm{d}\prior}\right)} \prior\bk{D}.
\end{equation}
\begin{proof}
Consider $f(x) = \exp(x-1) -1$ on $\R_+$, and $f(x) = \infty$ on $\R_-^*$. This results in $f^*(t) = t\log(t) + 1$ for all $t> 1/e$, and $1- \frac{1}{e}$ for $t < \frac{1}{e}$. This function can be upper bounded by $t\log(t) + 1$ for all $t>0$. Considering $D\geq 0$, $c=0$ and $\lambda >0$, and using \Cref{eq:ch2:loose_bound} in conjugation with this upper bound, this results in
\begin{align*}
\pi\bk{D}&\leq \lambda\prior\bk{\lambda^{-1}D\log(\lambda^{-1}D) +1} + \lambda\prior\bk{\exp\pth{\frac{\mathrm{d}\pi}{\mathrm{d}\prior}-1}-1}\\
&\leq \prior\bk{D\log(D)} - \log(\lambda)\prior\bk{D} + \lambda\prior\bk{\exp\pth{\frac{\mathrm{d}\pi}{\mathrm{d}\prior}-1}}.
\end{align*}
This bound holds for all $\lambda>0$ and is minimised for $\lambda^* = \frac{\prior\bk{D}}{\prior\bk{\exp\pth{\frac{\mathrm{d}\pi}{\mathrm{d}\prior}-1}}}$, yielding
\begin{align*}
\pi\bk{D}&\leq \prior\bk{D\log(D)} -\prior\bk{D} \log(\prior\bk{D}) +  \log\pth{\prior\bk{\exp\pth{\frac{\mathrm{d}\pi}{\mathrm{d}\prior}}}}\prior\bk{D}.
\end{align*}
\end{proof}
Our second bound involves a custom made penalisation which forces the ratio of density $\frac{\mathrm{d}\pi}{\mathrm{d}\prior}$ to be upper bounded; that is to say, $\exists r_{\max}$, $\forall r > r_{\max}$, $f(r) = \infty$. To obtain tractable expressions, we construct a convex $f$ such that $f'(0) = -\infty$ and $f'(r_{\max}) = \infty$. An instance of such U shaped convex function is the lower half circle, resulting in 
\begin{align*}
f_{\text{U}}(x) =\begin{cases} 1 - \sqrt{1 - (1-x)^2} & x\in[0,2]\\
\infty&\textup{else}.\end{cases}
\end{align*}

By rescaling this function, we obtain for $r_{\max}>1$ the convex function $f_{r_{\max}}(x) = f_{\text{U}}\pth{\frac{2x}{r_{\max}}} - f_{\text{U}}\pth{\frac{2}{r_{\max}}}$. The Legendre transform of $f_{\text{U}}$ has closed formed expression
\begin{align*}
f_{\text{U}}^*(t) = t + \abs{t}\sqrt{\frac{t^2}{1+t^2}}-1 + \sqrt{\frac{1}{1+t^2}},
\end{align*}
which results in 
\begin{align*}
f_{r_{\max}}^*(t) = &
\frac{r_{\max}}{2} \pth{t + \abs{t}\sqrt{\frac{r_{\max}^2t^2}{4+r_{\max}^2t^2}} -1 -\sqrt{\frac{r_{\max}^2}{4+r_{\max}^2}}}\\
&+ 2\sqrt{\frac{1}{4+r_{\max}^2t^2}} - 2\sqrt{\frac{1}{4+r_{\max}^2}}.
\end{align*}
While somewhat involved, $f_{r_{\max}}^*(t)$ behaves as $r_{\max}t - C_{r_{\max}}$ for $t\rightarrow \infty$, and as $-C_{r_{\max}}$ for $t\rightarrow -\infty$. The asymptotic for large values recaptures the non-penalized change-of-measure for $\frac{\mathrm{d}\pi}{\mathrm{d}\prior}\leq r_{\max}$ and $D\geq 0$, $\pi[D] \leq r_{\max}\prior\bk{D}$. The penalized bound improves on this behaviour by adding some flexibility. Moreover, since ${f_{r_{\max}}^*}'(\infty) = r_{\max}$, and $0\geq {f^*}'$, it follows that $\forall c, D$, $D+c$ is ${f^*}'$ integrable, and moreover, for all $D$ such that $\prior\bk{D>\infty}>0$, $\exists c, \prior\bk{{f_{r_{\max}}^*}'(D+c)} = 1$. Hence we can apply the second statement of \Cref{thm:ch2_1:theorem_optim_c}, and guarantee that minimising our bound in $c$ recovers the true $f$ divergence.

\section{Perspectives}
\label{sec:ch2:discussion}
\subsection{Change-of-measure with very weak penalisation}
Csizár-Donsker-Varadhan's change-of-measure \Cref{eq:ch2_1:change_measure} implies that the generalisation gap must have exponential moments to provide non trivial bounds. This condition is looser than the classic PAC-Bayes assumption that the risk is bounded. This raises the question of whether more efficient PAC-Bayes bounds could be built for looser penalisation than KL, leading to a strict bounded risk requirement. Obtaining such competitive bound would necessitate carefully designing "slow" convex function with tractable Legendre transform.

\subsection{Legendre transform of the entropy}
For some choices of $f$, it might be convenient to trade-off some tightness on the bound for more tractable expressions. A possible way to gain tractability could be to study the Legendre transform of the $f$-entropy, which is defined as
\begin{equation*}
\mathcal{E}_{f,\prior}: P \to \prior \bk{f\circ P}- f\left(\prior\bk{D}\right).
\end{equation*}
The $f$-entropy collapses to the $f$-divergence between $\pi$ and $\prior$ when evaluated for $P=\frac{\mathrm{d}\pi}{\mathrm{d}\prior}$, since $f(1)=0$. While the $f$-entropy might not be convex, an upper bound of $\mathcal{E}_{f,\prior}^*$ still results in an upper bound of $\mathcal{D}_{f,\prior}^*$. More generally, any extension of the $f$-divergence to a larger space can be used to upper bound $\mathcal{D}_{f,\pi}^*$.

\subsection{Change-of-measure inequalities for Variational PAC-Bayes}
Variational PAC-Bayes strategies construct posterior distribution belonging to a parametric family of probability measures. The definition of the Legendre's transform of the $f$-divergence, on the other hand, involves a worst case analysis performed on \emph{all} probability measures. Modifying the $f$-divergence to return $\infty$ outside of the variational family results in a decrease of the Legendre transform of this operator, leading to tighter bounds. Whether this decrease is significant or not would conceivably depend on the form of the variational family considered. Whether tractable expressions of the Legendre transform can be obtained remains uncertain. We expect the analysis to be more involved, and to depend on the form of the variational family. Moreover, the modified $f$-divergence might no longer be a convex operator if the exponential family is not a convex set (\emph{e.g.} exponential families are usually not convex sets).

In a similar spirit, tighter change-of-measure inequalities can be constructed by considering other forms of constraints. For instance, one may limit the search to the pruned posterior considered in \cite{mcallester1999} (\emph{i.e.} distributions such that $\frac{\dd \pi}{\dd\prior}(\gamma) < 1 \implies \frac{\dd\pi}{\dd\prior}(\gamma)=0$). Another option could be to only consider posterior distributions belonging to a $f$-divergence sphere (\emph{i.e.} disregarding all posterior distribution at distance more than $\tilde{r}$). For instance, considering Catoni's PAC-Bayes bound \Cref{eq:ch1:catoni} for a fixed temperature $\lambda$ in a bounded risk, all posteriors such that $\KL(\pi, \prior)\geq \lambda^{-1}$ result in a vacuous bound, and can therefore be disregarded. The same questions and limitations on the potential improvement and tractability occur.

%\textcolor{red}{BG: might be interesting to link to Jeremias Knoblauch's work on generalised Bayes}

\subsection{Finding approximately optimal \ensuremath{\lambda} and \ensuremath{c}}
\label{sec:ch2_1:lambda_selection}
As discussed in \Cref{sec:ch2_1:fdiv}, an appropriate choice of $c$ and $\lambda$ is necessary to obtain tight inequalities. In most cases, we could not explicitly compute which values are optimal, especially for $c$. Getting some theoretical or practical insight on how to pick these parameters in such a way as to obtain nearly optimal bounds is an exciting future avenue.

To approximate the optimal $c$, a strategy can consist in considering an idealized case where the generalisation gap takes a single, known value. For instance, for $D = \trisk - \risk$, the generalisation gap at a given predictor is a random variable of mean $0$, and if the number of observations $n$ is high, should have variations of order $O(n^{-\frac{1}{2}})$. Replacing $D$ by $0$ transforms the intractable renormalisation equation $\prior\bk{{f^*}'(D-c)} = 1$ by ${f^*}'(-\tilde{c}) = 1$ which has solution $\tilde{c} = - f'(1)$ (or more generally $c = \overline{D} - f'(1)$ when $D$ fluctuations close to $\overline{D}$).

\section{Conclusion}
PAC-Bayes generalisation relies on change-of-measure inequalities to transfer a concentration inequality on a fixed probability measure to all probability measures simultaneously. Additive change-of-measure inequalities can naturally be interpreted as Legendre transform of a penalisation term. In this section, we have studied how these Legendre transform can be upper bounded for a generic class of penalisation, $f$-divergence, which extends on the classic KL penalisation. The analysis shows a trade-off between the penalisation considered, and the assumptions which will be required on the risk. Weaker penalisation, which allows constructing posterior distribution further away from the prior, is paid for by stronger moment on the generalisation gap, and hence stronger assumptions on the risk.

Computing the exact Legendre transform of the penalisation involves optimisation on a single degree of freedom. This optimisation has no closed form expression in the general case, involving an intractable renormalisation condition. This makes the construction of tight PAC-Bayes bound with a generic $f$-divergence change-of-measure difficult. In this respect, the classic KL penalisation represents a sweet spot. First, the exact Legendre transform has a closed form expression as the renormalisation condition is tractable. Second, it involves exponential moment of the generalisation gap, matching the form used in Chernov's concentration inequalities. Finally, closed form expressions are available for the computation of the KL divergence for popular family of distributions such as Gaussian, facilitating the computation of the bound and its derivative for Variational PAC-Bayes settings.

\begin{acknowledgement}
A.P. acknowledges support by ANRT CIFRE grant 2021/1894. B.G. acknowledges partial support by the U.S. Army Research Laboratory and the U.S. Army Research Office, and by the U.K. Ministry of Defence and the U.K. Engineering and Physical Sciences Research Council (EPSRC) under grant number EP/R013616/1. B.G. acknowledges partial support from the French National Agency for Research, through grants ANR-18-CE40-0016-01 and ANR-18- CE23-0015-02, and through the programme “France 2030” and PEPR IA on grant SHARP ANR-23-PEIA-0008.
\end{acknowledgement}
\ethics{Competing Interests}{
The authors have no conflicts of interest to declare that are relevant to the content of this chapter.}
% \eject

%%%%%%%%%%%%%%%%%%%%%%%% referenc.tex %%%%%%%%%%%%%%%%%%%%%%%%%%%%%%
% sample references
% %
% Use this file as a template for your own input.
%
%%%%%%%%%%%%%%%%%%%%%%%% Springer-Verlag %%%%%%%%%%%%%%%%%%%%%%%%%%
%
% BibTeX users please use
% \bibliographystyle{}
% \bibliography{}
%
%\biblstarthook{
\bibliographystyle{abbrvnat}
\bibliography{bibliography}

\end{document}